\documentclass{ifacconf_amx}

\usepackage{graphicx}
\usepackage{svg}
\usepackage{amsmath,amssymb,amsfonts}
\usepackage{cite}
\usepackage{float}

\usepackage{siunitx}
\usepackage{fancyhdr}

\usepackage{lipsum}

\usepackage{tabularx}
\usepackage{booktabs}
\usepackage{xcolor}
\usepackage{multirow}

\usepackage[scaled]{helvet}

\usepackage[T1]{fontenc}

\newcommand{\chair}{Technical University of Munich\\Chair of Agrimechatronics}
\newcommand{\amxadress}{Technical University of Munich, Germany; Chair of Agrimechatronics; \\ Munich Institute of Robotics and Machine Intelligence (MIRMI)}

\makeatletter

\newcommand{\AuthorList}{}
\let\AuthorHeader\empty

\newcommand{\AddAuthor}[3][]{%

	\g@addto@macro\AuthorList{%
		\author[#3]{#2}%
	}

	\def\@tempname{#1}
	\ifx\@tempname\@empty
		\def\@tempname{#2}
	\fi

	\ifx\AuthorHeader\empty
		\xdef\AuthorHeader{\@tempname}%
	\else
		\xdef\AuthorHeader{\AuthorHeader, \@tempname}%
	\fi
}

\let\old@ssect\@ssect
\makeatother

\usepackage{hyperref}
\hypersetup{hidelinks}
\usepackage{cleveref}

\makeatletter
\def\@ssect#1#2#3#4#5#6{%
	\NR@gettitle{#6}%
	\old@ssect{#1}{#2}{#3}{#4}{#5}{#6}%
}
\makeatother

\graphicspath{{figures/}}
\usepackage{tikz}
\usetikzlibrary{positioning,arrows.meta,calc}
\definecolor{colA}{HTML}{2A78D6}
\definecolor{colB}{HTML}{EB6834}

\newcommand{\ISBN}{978-3-911430-18-0}
\newcommand{\PaperTitle}{INVAR-Agri: Detecting Discrepancies in Multilingual Operator Manuals of Agricultural Implements with a Fine-Tuned LLM}
\newcommand{\PaperDate}{26.10.2026}
\newcommand{\PaperKeywords}{machine translation, back-translation, translation checking, fine-tuning, operator manuals, agricultural machinery}

\AddAuthor[Oksanen T.]{Timo Oksanen}{}

\hypersetup{
    pdfinfo={
        Title={\PaperTitle},
        Author={\AuthorHeader},
        Keywords={\PaperKeywords},
        Publisher={Technical University of Munich},
        ISBN={\ISBN}
    }
}

\begin{document}
\begin{frontmatter}

   \title{\PaperTitle}

   \AuthorList

   \address{\amxadress}

   \begin{abstract}
   Operator manuals of agricultural implements are delivered in many languages, and the standards require that competent persons review every translation, which makes this review the bottleneck. This paper presents a proof of concept that compares two training workflows on the same LLM, EuroLLM-1.7B-Instruct, trained on operator manuals of one manufacturer. In approach A, a back-translator translates each language version back into English, and an English reader compares it with the English version. In approach B, a checker reads the English version and the translation together, where the checker reports differences. The test sets contain errors planted in numbers, units, opposite words, obligations, negations, and conditions in 24 languages, and they include unmodified text and meaning-preserving edits for alerts and false alarms. The trained back-translator concealed many planted errors, most of all weakened obligations, while translation-quality scores rose. In contrast, the separately trained checker detected most planted errors, although it was weak for units and for number changes unlike those in training. A check by the author confirmed the concealment. In addition, a simple rule check caught the number and unit changes, but it alerted on harmless rewrites. A check of alerts on real Finnish translations showed that most alerts were false, caused by language-specific wording, so the precision on real text remains a limitation. Within this setting, a trained checker might support, not replace, the review by competent persons, and firmer conclusions need more data.

   \end{abstract}


\end{frontmatter}

\section{Introduction}
\subsection{Translated operator manuals of agricultural implements}\label{sec:manuals}

A manufacturer of agricultural implements produces an operator manual for every machine type, which is delivered in the languages of the markets where the machine is sold. In the corpus of this study, each operator manual exists in 16 to 26 languages. The operator manual provides instructions for coupling, adjustment, field operation, transport on public roads, and maintenance. Safety-relevant content sits in small details, for example, tightening torques, speed limits, obligations such as ``must'', prohibitions such as ``do not'', and conditions such as ``when the engine is running''. One wrong number, unit, or word in a translation means the operator in that country receives a different instruction than the English version gives. Such a different instruction creates a risk of injury or machine damage. Every language version of an operator manual carries the same responsibility as the English version. In an earlier study, locally run LLMs answered operator questions from six language versions of one operator manual, where the answers on the non-English versions were less accurate than on the English version~\cite{Oksanen.2026}. It concluded that native-language manuals are preferable to relying on the translation abilities of the LLM for safety-critical use.

Machine translation and large language models (LLM) can translate a complete operator manual into many languages within hours at low cost. However, the check of each translation still needs a reviewer who reads that language and understands the machine. For example, an operator manual in 20 languages may require 20 reviewers, with one reviewer per language. The manufacturer rarely has such reviewers for every language and often relies on the translation provider. Consequently, checking, and not translating, is the bottleneck of multilingual operator manuals. The practical aim is that a technical expert reads English and checks all language versions with the support of an LLM, which is a proposed application that this paper does not test. However, this paper tests a precondition of such support, namely whether automatic checks keep visible or detect discrepancies in translated instructions.

\subsection{Two ways to let one expert check all languages}\label{sec:twoways}

In this paper, there are two ways to let one technical expert who reads English check every language version of an operator manual. Our approach A uses back-translation, where an LLM translates each language version back into English and the expert compares the back-translation with the English version. Users of machine translation, such as physicians, already rely on back-translation~\cite{Mehandru.2023}. However, a back-translation can repair an error on its way back, so that the error disappears~\cite{Somers.2005}. In contrast, approach B uses direct checking, where an LLM reads the English version and the translation together and reports what differs. The same idea underlies quality estimation and the detection of critical errors in machine translation~\cite{Specia.2021,Guerreiro.2024}. Both approaches depend on an LLM, which can be trained on the manufacturer's own operator manuals. This paper investigates how training each workflow changes whether planted discrepancies stay visible in approach A or are detected in approach B, how often each workflow alerts on unmodified text and on meaning-preserving edits, and how detection transfers to text not copied from training manuals and to changes unlike those in training.

Such an LLM supports, not replaces, the review by competent persons required by the standards~\cite{IEC.2019}. The check of an English reader could direct that bilingual review, but it cannot take its place. In addition, this paper tests the hypothesis concerning hiding in text close to the training text, as stated in Section~\ref{sec:measures}. For this study, EuroLLM-1.7B-Instruct is used for both approaches~\cite{Martins.2024}. This choice follows the preferred direction of the project, which is a European model for a study on European languages. In addition, the question of which model performs best is not a research question of this paper.

\subsection{Contribution}\label{sec:contribution}

This paper compares two training workflows on the same LLM, the same operator manuals, and the same test items, where each is evaluated in untuned and trained states. The two workflows differ beyond their task, as the back-translator learns from all translation pairs to reproduce the English version, selecting its best pass by chrF, whereas the checker learns from filtered correct pairs and synthetic errors to report differences, selecting its best pass by balanced accuracy. The paper builds its own test sets, called INVAR-Agri, which contain planted errors in 24 languages from public operator manuals of agricultural implements. The trained back-translator of approach A concealed planted errors that its untuned version showed, whereas the trained checker of approach B detected most of them. Manufacturers can learn several lessons from this comparison, for example, that a back-translator trained on the manufacturer's own manuals can make a review by back-translation less safe even when translation-quality scores improve. In addition, a checker trained with synthetic errors from the same manuals detects most planted errors, but not every unit change. A deterministic check of numbers and units remains useful beside the checker.

The paper represents a proof of concept within operator manuals of agricultural machinery, rather than a benchmark. As a result, the paper makes no claims beyond this specific scope. In addition, the paper reports the direction and the size of the effects instead of exact performance figures.

\section{State of the Art}
\subsection{Back-translation and round-trip checks}\label{sec:sota-bt}

Round-trip translation translates a text into another language and back, which allows a reviewer to compare the result with the original. Early work showed that a bad forward translation can come back perfect, for example with the idiom ``tit for tat''~\cite{Somers.2005}. In addition, the same work found almost no correlation between the BLEU scores of the forward translation and of the round-trip translation~\cite{Somers.2005}. With semantic measures, round-trip scores correlated well with human ratings at the level of whole systems, whereas they stayed weak at the level of single sentences~\cite{Moon.2020}. However, the backward system can recover the content of the source and make a wrong translation look correct~\cite{Agrawal.2022}.

Users of machine translation who cannot read the target language use back-translation as an aid. In a study with 52 users, back-translation raised their confidence but not the quality of their translations~\cite{Zouhar.2021}. In addition, back-translation caught a critical error in a high-risk medication instruction more often than quality estimation did in a study with physicians~\cite{Mehandru.2023}. In a user study, monolingual English users decided whether English-to-Spanish translations could be shared, and users received one of four kinds of feedback or none~\cite{Ki.2025}. Back-translation from a general system raised the accuracy of their decisions the most and led to the least over-reliance, whereas LLM explanations, as explicit feedback, caused more changes of decisions and more over-reliance. Nevertheless, in a pilot study, users trusted fluent but wrong output more than disfluent output~\cite{Martindale.2018}. These studies use general translation systems, and none of them tests a back-translator fine-tuned on the same documents it later checks.

\subsection{Quality estimation and critical error detection}\label{sec:sota-qe}

COMET scores a translation from the source, the translation, and a reference using a multilingual encoder~\cite{Rei.2020}. In addition, quality estimation scores a translation without a reference, based on the source and the translation alone. The WMT 2021 shared task added the detection of critical errors, which are deviations in meaning with health, safety, legal, or financial consequences, for example, a removed negation or a wrong number~\cite{Specia.2021}. The best system reached a Matthews correlation of only 0.55 for English to German in this task~\cite{Specia.2021}. Moreover, xCOMET marks error spans with their severity and also works without a reference~\cite{Guerreiro.2024}.

Several systems train with planted synthetic errors instead of new human labels. Labels made from parallel data alone are about as good as human labels for quality estimation~\cite{Tuan.2021}. For example, SMAUG plants errors in numbers, named entities, and negation, among others~\cite{Alves.2022}. Rule-based detectors find errors in numbers and units with high precision~\cite{Raunak.2022}. However, even a strong quality estimation model missed removed negations and antonyms in an adversarial test~\cite{Kanojia.2021}. Approach B follows this recipe of planted errors, using the same LLM as approach A and errors planted in operator manuals.

\subsection{Fine-tuning LLMs on multi-parallel domain data}\label{sec:sota-ft}

Fine-tuning on data of one domain adapts a translation model to the domain, although overfitting is common when the fine-tuning data is small or repetitive~\cite{Saunders.2022}. For LLaMA-2-7B, translation quality peaked with 10 to 100 thousand parallel pairs, whereas millions of pairs appeared to wash out what the model already knew~\cite{Xu.2024}. A company setting required at least 5000 examples for fine-tuning on translation memories~\cite{Vieira.2024}. Multi-parallel data consists of one sentence translated into many languages. A model trained from many languages into English likely overfitted the English side of multi-parallel data~\cite{Aharoni.2019}. A recent study named this effect directional degeneration, which results in fluent but less faithful output in the pivot language~\cite{Luo.2026}. In addition, the repetition of a target leads to memorisation and to copies of training targets for unseen sources~\cite{Raunak.2022a,Carlini.2023}.

EuroLLM is a family of open multilingual models for the European languages, which contains versions with 1.7, 9 and 22 billion parameters~\cite{Martins.2024,Martins.2025,Ramos.2026}. Studies that fine-tuned EuroLLM concern at most eight language pairs; for example, EuroLLM-1.7B-Instruct was fine-tuned for English to Spanish terminology~\cite{Zhu.2025}. Moreover, EuroLLM-9B was fine-tuned for English to Japanese~\cite{Zhang.2025} and for eight pairs from English~\cite{Guttmann.2025}. None of these studies fine-tunes EuroLLM on operator manuals in many languages or measures whether fine-tuning hides errors.

\subsection{Technical documentation of machinery}\label{sec:sota-docs}

The Machinery Directive requires instructions in the official language or languages of the Member State, and marks versions as ``Original instructions'' or ``Translation of the original instructions''~\cite{EU.2006}. In addition, the Machinery Regulation, which is applicable from 2027, requires instructions in a language that users easily understand, as determined by the Member State, and allows digital instructions~\cite{EU.2023}. ISO 3600 serves as the standard for operator manuals of tractors and agricultural machinery, requiring explained signal words and warnings before the instructions they concern, and recommending manuals in the appropriate language for each country of sale and SI units~\cite{ISO.2022}. Moreover, ISO 20607 asks for text that is easily translatable and for instructions without weakening words and without negative words~\cite{ISO.2019}. Its draft revision keeps the clause on languages unchanged and adds fully digital instruction handbooks~\cite{ISO.2025}.

IEC/IEEE 82079-1 recommends the review of translations for technical and linguistic accuracy by competent persons~\cite{IEC.2019}. These competent persons are translators who are fluent in both languages, native in the target language, and familiar with the product~\cite{IEC.2019}. None of these documents addresses machine translation or an automatic check of translations. As a result, this review by competent persons constitutes the bottleneck that this paper addresses.

\section{Materials}
\subsection{Corpus of operator manuals}\label{sec:corpus}

The corpus contains 12 operator manuals of agricultural implements from one manufacturer, the Kverneland Group, covering 10 implement types, for example, rotary rakes, seed drills, a mower conditioner, a trailed sprayer and a fertiliser spreader~\cite{KV.CultivatorEnduro,KV.MowerKV43xx,KV.PlanterOptimaR,KV.PowerHarrowNGM,KV.QualidiscPro,KV.RakeKV9439,KV.RakeKV9476C,KV.SeederDA,KV.SeederUdrill,KV.SprayerIXtrackT3,KV.SpreaderTLX,KV.TedderKV8576}. Every operator manual exists in 16 to 26 language versions, which include English. Since EuroLLM-1.7B-Instruct does not cover Serbian, the study uses English and the 24 other languages that EuroLLM-1.7B-Instruct covers. In addition, the corpus aligns each English text unit, which may be a sentence, a paragraph or a list, with the same unit in every other language version. The manufacturer publishes these operator manuals on its public portal. 

The corpus was built from the digital PDF files of the public portal, where each file was recorded with its revision and checksum. The manuals share one page layout across languages, whereas single translations insert or omit pages. Therefore, every language version was first mapped page by page to the English version, using fingerprints of the images, drawings and numbers on each page, which matched at least 98.3~\% of the English content pages in every manual and language. A layout model (Docling) separated prose from tables, labels, page headers and other noise, while the text was taken from the text layer of the PDF files. Within each page, text blocks were aligned to the English blocks by position, type, reading order, length and numbers, and a median of 96.1~\% of the English blocks were aligned. Two judge models, Gemma 4 (26B) and Clef, calibrated on 60 Finnish pairs reviewed by the author, and a review by the author removed 3735 misaligned pairs, which is about 1~\%, and flagged 2134 pairs. The judges scored one pair per manual, page, and language, and they also scored every pair with a high alignment cost, which totaled about 63000 pairs. In addition, the flagged pairs were kept in the corpus but left out of all training, development, and test data, as were all list units whose language the corpus does not verify. In addition, one manual for a plough was left out because its text was reflowed in every language. The corpus totals 19272 English blocks and 354390 aligned block pairs.

A new operator manual takes much of its text from older manuals of the same machine type, which is normal practice in technical documentation and not a weakness. In the two test manuals, 69~\% and 42~\% of the text units appear word for word in the training manuals. The study compares every test unit with its most similar English unit of the training manuals, using Jaccard similarity of character 5-grams of the normalised English text, where candidates are found by MinHash. In addition, the data is divided into four groups: exact copies, high similarity of at least 0.8, partial similarity from 0.5 to 0.8, and new text below 0.5. The study builds one training pair per text unit and language, which results in 198030 training pairs after removing 63362 exact duplicates with the same language, translation, and English text. In addition, a development pair identical to a training pair is dropped from the development data, whereas similar standard text remains in both. Moreover, the study trains on 10 manuals and keeps 10~\% of their pages, totaling 134 pages, aside for development. The study tests on 2 manuals of known machine types, specifically the KV 9439/9443 rotary rakes from 2012 with 20 languages~\cite{KV.RakeKV9439} and the DA seed drills from 2021 with 22 languages~\cite{KV.SeederDA}. The two test manuals complement each other in machine type and in language subset. The test sets cover complete sentences and paragraphs only. Headings, tables, captions, and text in figures are not covered. The study treats the English version as the reference for every check. However, the source language and the translation path of each manual are not known.

\subsection{Model and helper models}\label{sec:models}

Both approach A and approach B use the same model, EuroLLM-1.7B-Instruct, which is an open-weight multilingual model trained for the official languages of the European Union and further languages, for example, Norwegian, Russian and Ukrainian~\cite{Martins.2024}. The study fine-tunes all weights of EuroLLM-1.7B-Instruct on one NVIDIA DGX Spark with 128 GB of unified memory. A training run needs about 65 GB of memory. The memory of one DGX Spark is enough for full fine-tuning of a model with 1.7 billion parameters. Therefore, tuning with low-rank adapters (LoRA) is not needed for this model. This memory decided the size of the model in this study.

Three further models prepare and validate the data. Gemma 4 (26B) plants the word-level errors of the test sets, whereas gpt-oss-120b plants the word-level errors of the training data of the checker. The study uses a different planter for the training data on purpose, so the checker cannot simply learn the habits of one planter. Clef, a decision model released by Cloudflare in 2026, returns a probability for each allowed answer instead of free text~\cite{Cloudflare.2026}. This model builds on Qwen3.8-27B and scores all answers of all questions in one pass~\cite{Cloudflare.2026a}. An independent study found Clef consistent and cheap, but biased in 5 of 12 tests for judge bias~\cite{Lukassen.2026}. Clef performs two different tasks in this study. As validator of a planted error, Clef receives the English version, the original translation, the changed translation and the intended change. It accepts a planting at a probability of the intended change of at least 0.5, which was set on the Finnish check (Section~\ref{sec:testsets}). As an alternative judge in approach A, Clef receives only the English version and the back-translation, where it misses weakened obligations (Section~\ref{sec:hides}). In addition, a further Finnish check of 10 accepted and 30 rejected plantings (Section~\ref{sec:checks}) showed all 10 accepted plantings valid, whereas 26 of the 30 rejected ones were valid as well. In the audited Finnish examples, the validation was precise but strict, as it favoured short changes in one place. This occurred because Clef scored even clear opposites low, the length rule removed long conditions, and the one-span rule removed Finnish modal forms that need two changes. All three models are run through Ollama on the same machine.

Clef follows the concept of decision models that Jev of TypeSafe AI introduced, and Clef is compatible with the interface of Jev~\cite{Cloudflare.2026}. This study applied Clef on the local server a few days after its release, as a judge of the corpus alignment (Section~\ref{sec:corpus}). On the 60 Finnish pairs reviewed by the author, Clef achieved an accuracy of 0.983, whereas Gemma 4 (26B) reached 0.917 and gpt-oss-120b reached 0.850 and needed about 16 times the time per pair of Gemma 4 (26B). Clef processed about 63000 pairs at about 0.6 s per pair, without failures. Moreover, the design of the questions mattered, where two separate yes/no questions, whether the text is the same passage and whether a part is missing, worked best.

\subsection{Test sets with planted errors}\label{sec:testsets}

The study builds the test sets, called INVAR-Agri, from the two test manuals, where the name refers to the invariants of a translation, that is, the content that must not change; each test item consists of one translated text unit with exactly one planted error and the English version of the unit stays unchanged. The complete list of six error types consists of number, unit, polarity, obligation, negation, and condition, with one example of each presented in Table~\ref{tab:errortypes}. Polarity replaces a word by its opposite, for example, raise and lower, or on and off, while obligation weakens ``must'' to ``should'' or ``may''. Negation removes a negation, and condition removes a condition, such as ``after cleaning''. In addition, the six error types touch the elements that ISO 20607 asks drafters to control, including words that weaken an instruction, negative words, and consistent units~\cite{ISO.2019}. Each error type possesses a different consequence for the operator, as the examples of Table~\ref{tab:errortypes} show, for example, a changed number turns a service interval of 30 hours into 300 hours. In addition, a changed unit turns a tightening torque of 85 Nm into 85 N, an opposite word turns ``switch off the PTO shaft drive'' into ``switch on'', and a removed negation turns ``must not be connected'' into ``may be connected'' for the electrical supply during fitting. A weakened obligation turns a required machine weight into a recommendation, while a removed condition makes an action unconditional. The study does not weight the error types by risk, so pooled detection rates do not express equal safety importance of the types.

Rules change numbers and units by multiplying or dividing a number by ten, changing one digit, or swapping a unit, for example, mm and cm or Nm and N. Gemma 4 (26B) plants the four word-level types, which are filtered by automatic checks that reject a planting with more than one changed span, changed numbers, or an unusual change of length. Clef accepts a planting only if the probability of the intended change is at least 0.5, and Gemma 4 (26B) plants conditions only in simple sentences with one main clause and one condition. The study planted the word-level types first in 8 languages: German, French, Finnish, Spanish, Italian, Polish, Estonian, and Lithuanian, and later in the other 16 languages as well. A check of the Finnish plantings found 34 of 34 correct after fixes of the planting rules were applied. Claude (Opus 5.5) checked a sample of 192 plantings in the 16 added languages, where it found 185 valid and 7 (3.6~\%) invalid, including 0 of 48 polarity, 3 of 48 obligation, 3 of 48 negation and 1 of 48 condition errors. The accepted word-level items per type and language ranged from 3 for obligation in Bulgarian to the limit of 50.

The test sets contain 5332 planted errors in 24 languages, and Table~\ref{tab:errortypes} gives the counts per type. The unmodified translations of the test manuals comprise 18317 text units, which serve to measure alerts. The comparison of the two approaches uses the same sample of 2016 unmodified units for every system, stratified by language. In addition, 432 edits that keep the meaning serve to measure false alarms on harmless changes, for example, a decimal comma instead of a point, a thousands separator, or ``N·m'' instead of ``Nm''.

\begin{table*}[t]
\centering
\caption{The six error types of the test sets, the planting method, and the number of test items in 24 languages, including one example per type from the Finnish version of the rake manual with the planted or removed words in bold.}
\label{tab:errortypes}
\small
\begin{tabularx}{\textwidth}{llr>{\raggedright\arraybackslash}X>{\raggedright\arraybackslash}X}
\toprule
Error type & Planted by & Items & English version & Finnish version, planted or removed words in bold \\
\midrule
number & rule & \num{1094} & after 30 hours of operation. & 30 käyttötunnin jälkeen. $\rightarrow$ \textbf{300} käyttötunnin jälkeen. \\
unit & rule & \num{383} & Retighten the bolts to a tightening torque of 85~Nm. & Kiristä ruuvit kiristystiukkuuteen 85~Nm. $\rightarrow$ Kiristä ruuvit kiristystiukkuuteen 85~\textbf{N}. \\
polarity & LLM & \num{1098} & Switch off the PTO shaft drive. & Kytke nivelakselin käyttö pois päältä. $\rightarrow$ Kytke nivelakselin käyttö \textbf{päälle}. \\
obligation & LLM & \num{756} & The empty weight of the tractor must be greater than the weight of the machine. & Traktorin omapainon on oltava koneen painoa suurempi. $\rightarrow$ Traktorin omapainon \textbf{tulisi olla} koneen painoa suurempi. \\
negation & LLM & \num{988} & The electrical supply to the tractor must not be connected when the lighting equipment is being fitted. & Asentaessa valaistusjärjestelmää paikalleen traktorin virransyöttö ei saa olla liitettynä. $\rightarrow$ Asentaessa valaistusjärjestelmää paikalleen traktorin virransyöttö \textbf{saa} olla liitettynä. \\
condition & LLM & \num{1013} & Lubricate all bearings after cleaning. & Voitele kaikki laakerit \textbf{puhdistuksen jälkeen}. $\rightarrow$ Voitele kaikki laakerit. \\
\bottomrule
\end{tabularx}
\end{table*}

\section{Methods}
\subsection{Approach A: back-translation review}\label{sec:approachA}

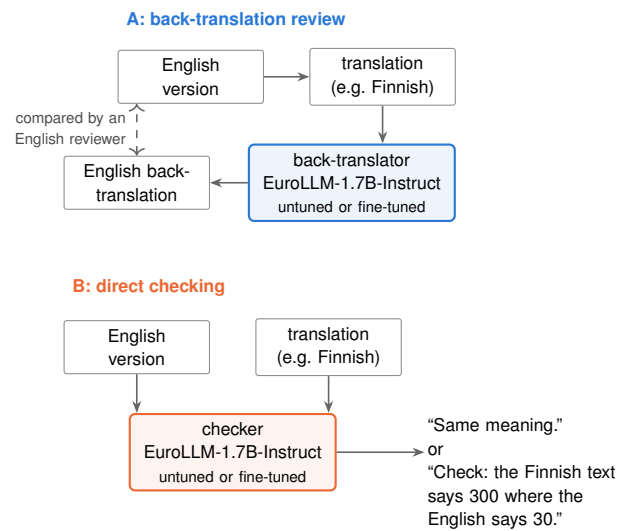
\begin{figure}[b]
\centering
\begin{tikzpicture}[
  font=\sffamily\scriptsize,
  doc/.style={draw=black!45, rounded corners=1pt, align=center, minimum height=7mm, text width=17mm, fill=white},
  llmA/.style={draw=colA, line width=0.8pt, rounded corners=2pt, align=center, minimum height=9mm, text width=25mm, fill=colA!8},
  llmB/.style={draw=colB, line width=0.8pt, rounded corners=2pt, align=center, minimum height=9mm, text width=25mm, fill=colB!8},
  answer/.style={align=left, text width=27mm, inner sep=1pt},
  note/.style={align=right, text width=17mm, inner sep=1pt, font=\sffamily\tiny, text=black!70},
  arr/.style={-{Stealth[length=1.6mm]}, black!60, line width=0.6pt},
  lab/.style={font=\sffamily\scriptsize\bfseries, anchor=west}
]
\node[lab, text=colA] (ta) at (0,0) {A: back-translation review};
\node[doc, below=2mm of ta.south west, anchor=north west] (enA) {English\\version};
\node[doc, right=6mm of enA] (xA) {translation\\(e.g.\ Finnish)};
\node[llmA, below=5mm of xA.south east, anchor=north east] (bt) {back-translator\\EuroLLM-1.7B-Instruct\\{\tiny untuned or fine-tuned}};
\node[doc, left=5mm of bt] (btout) {English back-\\translation};
\draw[arr] (enA) -- (xA);
\draw[arr] (xA.south) -- (xA.south |- bt.north);
\draw[arr] (bt) -- (btout);
\draw[arr, dashed, <->] (btout.north) -- node[note, left=1mm] {compared by an English reviewer} (btout.north |- enA.south);
\node[lab, text=colB, below=8mm of btout.south west, anchor=north west] (tb) {B: direct checking};
\node[doc, below=2mm of tb.south west, anchor=north west] (enB) {English\\version};
\node[doc, right=6mm of enB] (xB) {translation\\(e.g.\ Finnish)};
\node[llmB, below=5mm of $(enB.south)!0.5!(xB.south)$, anchor=north] (ck) {checker\\EuroLLM-1.7B-Instruct\\{\tiny untuned or fine-tuned}};
\node[answer, right=3mm of xB.east, anchor=north west, yshift=-9mm] (ans) {``Same meaning.''\\or\\``Check: the Finnish text says 300 where the English says 30.''};
\draw[arr] (enB.south) -- (enB.south |- ck.north);
\draw[arr] (xB.south) -- (xB.south |- ck.north);
\draw[arr] (ck.east) -- (ck.east -| ans.west);
\end{tikzpicture}
\caption{Approach A uses the LLM to translate the translation back into English for comparison by an English reviewer, whereas approach B uses the LLM to read the English version and the translation together to answer ``Same meaning.'' or ``Check:'' with the difference, using the same untuned or fine-tuned LLM in both approaches.}
\label{fig:approaches}
\end{figure}

Our approach A uses EuroLLM-1.7B-Instruct to translate every language version back into English, where an English-speaking reviewer compares this back-translation with the English version (Fig.~\ref{fig:approaches}). The untuned and the fine-tuned back-translator use the same prompt and greedy decoding. This prompt contains a system instruction to translate exactly, including every number, unit, warning word, negation and condition, and the user message from the model card of EuroLLM-1.7B-Instruct. In addition, the study selected this prompt among ten prompt variants on development data only, by COMET~\cite{Rei.2020}. The prompt reads as follows:

\begin{quote}\footnotesize\textit{System:} You translate texts from operator manuals for agricultural machinery into English. Translate exactly what the text says, including every number, unit, warning word, negation and condition. Do not correct, complete or improve the content. Output only the English translation.\\
\textit{User:} Translate the following \textlangle language\textrangle{} source text to English: \textlangle language\textrangle: \textlangle text\textrangle{} English:\end{quote}

The study replaces the reviewer with an automatic decision per test item, which results in either a flag or no flag. Element rules compare the English back-translation with the English version, and they flag a planted error when the back-translation shows the planted value, for example 300 instead of 30 or ``should'' instead of ``must''. Each back-translation of a planted item results in one of three outcomes, which are carried (the planted value appears), hidden (the original value appears, restoring the instruction), and garbled (neither value appears). In addition, only a carried error counts as detected. The record of each planting holds the English side of the change, for example the English word and its opposite or the English modal that was weakened, and the element rules search the back-translation for these English forms. On unmodified units and meaning-preserving edits, the element rules flag any difference between the back-translation and the English version in numbers, units, modal class, negation or opposite words. In addition, a garbled back-translation is not counted as detected, as a reviewer cannot tell a corrupted sentence from an error. Counting garbled back-translations as visible would change the fall by training little, from 95~\% to 77~\% instead of from 87~\% to 67~\%, for the error types without conditions. For removed conditions, Clef judges whether the back-translation still states the condition. An alternative decision is also provided, where Clef judges whether the back-translation and the English version say the same, and a probability below 0.5 counts as a flag~\cite{Cloudflare.2026}.

The answers of the author's check described in Section~\ref{sec:checks} agreed more often with the element rules than with Clef. The results of approach A use the element rules, while the Clef decision serves as a comparison.

\subsection{Approach B: direct checking}\label{sec:approachB}

Our approach B uses the same model, EuroLLM-1.7B-Instruct, which reads the English version and the translation together and answers ``Same meaning.'' or ``Check:'' followed by a short description of the difference, as shown in Fig.~\ref{fig:approaches}. For example, the model may answer ``Check: the Finnish text says 300 where the English says 30.'' An answer that starts with ``Check:'' counts as a flag, whereas an answer in any other form counts as no flag. Answers outside the required form are reported separately, which allows for a lenient reading of them as a sensitivity analysis in Section~\ref{sec:learns}. In addition, this lenient reading follows a fixed rule without a person or a model, where an answer counts as a flag if it contains an English word for a difference, for example ``differ'', ``missing'' or ``instead of'', and no statement of the same meaning, for example ``same meaning'' or ``identical''. The prompt reads as follows:

\begin{quote}\footnotesize\textit{System:} You check translations of operator manuals for agricultural machinery. Compare the English original with the \textlangle language\textrangle{} translation. If the translation says the same, answer ``Same meaning.'' Otherwise answer ``Check:'' and say exactly what differs.\\
\textit{User:} English: \textlangle English text\textrangle{} \textlangle language\textrangle: \textlangle translation\textrangle\end{quote}

In addition, the untuned checker receives the same prompt with three examples as a second untuned variant, using one correct pair, one number error, and one obligation error from the training data. The prompt of the checker is used as written, without being selected among variants.

The training data of the checker comes only from the 10 training manuals and need no human annotation~\cite{Tuan.2021}. Correct pairs teach the answer ``Same meaning.'' However, the human translation may differ from the English version in numbers, units, obligation or negation. The study leaves out pairs whose untuned back-translation disagrees with the English version in these elements due to these differences. This filter left out 18.3~\% of the training pairs, and the 8263 correct examples were drawn from the remaining pairs. In addition, the training data contained 3000 planted number errors, 2524 unit errors, 975 polarity errors, 728 obligation errors, 843 negation errors and 962 condition errors. The four word-level types among them were accepted from 1250 candidates per type, using the same automatic checks and Clef threshold as the test sets. Pairs with one planted error teach the answer ``Check:'', where the explanation follows a fixed template filled from the record of the planting. Planted errors also appear in the training data of xCOMET and in the test set of the error generator SMAUG~\cite{Guerreiro.2024,Alves.2022}.

The training errors come from a different generator than the test errors on purpose, because the test measures whether the checker generalises to kinds of changes it has not seen. The training data doubles, halves or multiplies numbers by 1.5, while the test data multiplies or divides them by ten or changes one digit. The training data swaps the unit pairs mm and m, kg and t, bar and MPa, km/h and km, whereas the test data swaps other pairs. The training data is composed of 17295 examples in 24 languages, which include 9032 with an error and 8263 correct, and 2026 examples for development.

A rule check without a model serves as an engineering baseline, which flags a unit if the numbers or the pairs of number and unit differ between the English version and the translation. The normalisation of the rule check covers decimal separators and unit symbols in Cyrillic script only. In addition, it does not cover thousands groups separated by a space, which split one number into two, or the unit spelling N·m with a middle dot, so these meaning-preserving edits raise alerts. The rule check does not convert between units, for example 8.5 daNm for 85 Nm, and valid conversions of units are not included in the evaluation. The study evaluates the rule check alone and combined with the checker, where a unit is flagged if either flags it, similarly as in~\cite{Raunak.2022}.

\subsection{Fine-tuning}\label{sec:finetuning}

Both approach A and approach B use the same recipe, which consists of full fine-tuning of all weights of EuroLLM-1.7B-Instruct with a learning rate of \num{2e-5} and an effective batch size of 64 across 3 seeds. The two training workflows share the base model and these optimiser settings, whereas they differ in training data, supervision, filtering and development measure. Other studies fine-tuned EuroLLM models with LoRA~\cite{Zhu.2025,Guttmann.2025}. However, this study uses full fine-tuning on purpose, which makes no compromise in the method. Therefore, the results need no assumptions about how adapters behave. The learning rate was chosen in a short sweep on development data. In addition, approach A learns from 198030 training pairs to translate each language version into the English version, whereas approach B learns from 17295 examples to answer with the verdict and its explanation. One English sentence is the target of about 20 training pairs, with one pair per language, and such multi-parallel data invites memorisation of the English side~\cite{Aharoni.2019}. Due to the risk of memorisation, the study rotates the languages, where every pass shows each English text once with one language that changes from pass to pass.

\begin{figure}[t]
\centering
\includegraphics[width=\linewidth]{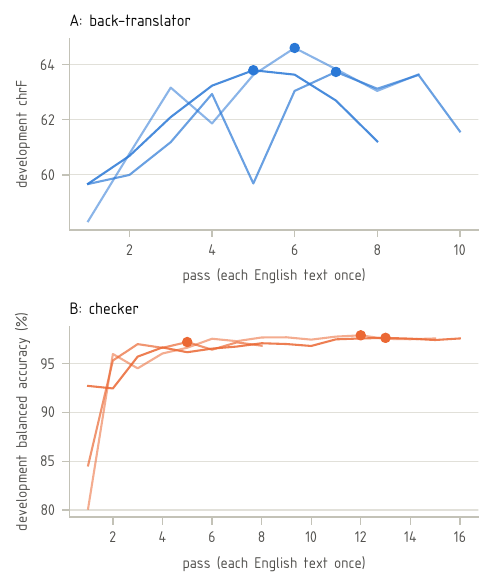}
\caption{Development score after every pass for three seeds per approach, where dots indicate the best pass kept as the model.}
\label{fig:devcurves}
\end{figure}

The study evaluates on the development data after every pass, keeps the best pass, and stops after 3 passes without improvement. The development measure is chrF for approach A~\cite{Popovic.2015} and balanced accuracy for approach B. In addition, selecting the pass of approach A by chrF against the English version rewards a back-translation that restores the English text, as a standard translation workflow would. Table~\ref{tab:training} and Fig.~\ref{fig:devcurves} give the details of every run, where the best pass lies between pass 5 and 7 for approach A and between pass 5 and 13 for approach B. The development chrF of approach A declined after its best pass, whereas the checker of approach B rose within two passes to a balanced accuracy above 95~\%, staying there until early stopping. However, an earlier run of approach A trained on all pairs for 2 epochs memorised the English targets, which resulted in a rise of development loss while chrF stayed flat. The study keeps the two models of this run, after epoch 1 and epoch 2, as an overtrained contrast for approach A.

\begin{table}[b]
\centering
\caption{Training of the back-translator for approach A and of the checker for approach B, including the values of seeds 1, 2, and 3.}
\label{tab:training}
\footnotesize
\begin{tabularx}{\linewidth}{>{\raggedright\arraybackslash}X l l}
\toprule
 & A: back-translator & B: checker \\
\midrule
training examples & \num{198030} pairs & \num{17295} examples \\
shown per pass (one per English text) & \num{10553} & \num{6947} \\
optimiser steps per pass & 165 & 109 \\
development measure & chrF & balanced accuracy \\
best pass, seeds 1 / 2 / 3 & 6 / 7 / 5 & 12 / 5 / 13 \\
passes run (early stopping) & 9 / 10 / 8 & 15 / 8 / 16 \\
best development score & 64.6 / 63.7 / 63.8 & 97.9 / 97.2 / 97.6 \\
training time (h) & 1.8 / 2.0 / 1.6 & 2.4 / 1.3 / 2.5 \\
peak GPU memory (GB) & 65 / 65 / 67 & 63 / 63 / 63 \\
\bottomrule
\end{tabularx}
\end{table}

\begin{table}[t]
\centering
\caption{Settings of the models, decoding, and training, as well as the denominators of the test sets and of the checks by the author.}
\label{tab:settings}
\footnotesize
\setlength{\tabcolsep}{3pt}
\begin{tabularx}{\linewidth}{>{\raggedright\arraybackslash}p{0.27\linewidth} >{\raggedright\arraybackslash}X}
\toprule
\multicolumn{2}{l}{\textit{Models}} \\
LLM of both approaches & utter-project/EuroLLM-1.7B-Instruct, revision a25c7fa \\
COMET & Unbabel/wmt22-comet-da, revision 2760a223 \\
helper models & Gemma 4 (26B), gpt-oss-120b, Clef; Ollama 0.35.1 \\
audit of plantings & Claude (Opus 5.5), 192 plantings in the 16 added languages \\
\midrule
\multicolumn{2}{l}{\textit{Decoding and training}} \\
decoding & greedy; at most 512 new tokens (back-translation), 128 (checker) \\
training & full fine-tuning, fused AdamW, learning rate \num{2e-5} with cosine decay and 3~\% warm-up, weight decay 0, gradient clipping 1.0, effective batch 64, bf16 with fp32 master weights, loss on the answer and its end marker only, one language per English text and pass, early stopping after 3 passes without improvement, seeds 1, 2 and 3 \\
checker answers & templates filled from the planting record, e.g. ``Check: the negation (`not') is missing in the Finnish text.'' \\
\midrule
\multicolumn{2}{l}{\textit{Denominators}} \\
planted errors & 5332: number 1094, unit 383, polarity 1098, obligation 756, negation 988, condition 1013 \\
unmodified units & 2016 (84 per language, random; the same sample for every system) \\
meaning-preserving edits & 432 (6 to 23 per language, where an edit was possible): decimal separator 274, thousands separator 111, unit spelling 47 \\
planting filter & one changed span, numbers kept, length ratio 0.7 to 1.3 (conditions 0.4 to 1.0), Clef probability of the intended change at least 0.5 \\
checks by the author & 117 back-translations (H1); 40 Finnish plantings; 40 unmodified Finnish units \\
\bottomrule
\end{tabularx}
\end{table}

\subsection{Measures}\label{sec:measures}

The study first evaluates both approach A and approach B in the standard way on the unmodified text of the two test manuals, as described in Section~\ref{sec:standard}. For approach A, the study computes COMET~\cite{Rei.2020} and chrF~\cite{Popovic.2015} of the back-translation against the English version. These metrics are computed separately for new text, which has no similar unit in the training manuals, and for text close to the training text. In addition, Clef judges whether the back-translation says the same as the English version, and the study counts word-for-word copies of the English version. For approach B, the study counts the flags on unmodified text, called alerts, and the answers in the required form.

Detection is measured as the share of planted errors that a system flags, per error type. Flags on unmodified units are called alerts, because an unmodified translation may itself differ from the English version. These alerts were measured on the same sample of 2016 units for every system. In addition, the flags on the 432 meaning-preserving edits, which are known to keep the meaning, are called false alarms. The change by fine-tuning is calculated as the mean of the 3 fine-tuned seeds minus the untuned model. The range of the three seeds shows the variation of training. In addition, 95~\% intervals from a bootstrap that resamples whole English texts show the uncertainty of the test sample, because one English text appears in up to 24 languages and on several pages. In this proof of concept, the direction and the size of the change matter, not the exact percentages. Before any result, the study pre-registered the hypothesis that fine-tuning raises the hiding of planted errors more in text close to the training text than in new text. A logistic model, one per fine-tuned seed against the untuned model, predicts whether a planted error is hidden based on training, the similarity group of the unit, the interaction of training and similarity group, the error type, and the language. The interaction of training and similarity group provides the test quantity, which is accompanied by 95~\% intervals from a bootstrap over whole pages. In this model, exact copies and units of high similarity with at least 0.8 together form the group of text close to the training text, as in Table~\ref{tab:headline}, while partial similarity forms a separate group and new text is the reference. In addition, the model covers the error types from number to negation, without conditions. The bootstrap over pages was a pre-registered choice, whereas the bootstrap over English texts was a sensitivity check (Section~\ref{sec:hides}). In addition, the protocols of the study were frozen before the results they concern, although they were not publicly registered. First, the hypothesis and this model were fixed before any back-translation of the test data; next, the comparison of the two workflows was framed after the first results of the checker, and the word-level error types were extended from 8 to 24 languages. Finally, the model was fitted after its descriptive rates were known, and all deviations were recorded as amendments of the protocols.

\subsection{Checks by the author and by LLM}\label{sec:checks}

The author checked the planted Finnish errors, as described in Section~\ref{sec:testsets}, and compared Finnish translations of the untuned back-translator during the choice of the prompt. In addition, the main check covers 117 back-translations, which include 10 per error type from the untuned and the fine-tuned back-translator using seed 1. Each item shows the English version, the English version with the planted error marked, and the back-translation. The question of each item asks which of the two English texts the back-translation matches. In this check, the sheet showed neither the system that produced a back-translation nor the automatic decisions. In 3 of 120 questions, the English version with the planted error made no sense, so the study left these 3 questions out. The English version with the planted error was built by rule from the record of the planting. In addition, the author judged all items of the untuned and of the fine-tuned back-translator, which were sampled independently. The 3 left-out questions included 2 negation and 1 obligation, and the 7 items that could not be judged were 4 polarity and one each of obligation, negation and condition.

Opus 5.5 was used to check the plantings of the 16 added languages on a sample (Section~\ref{sec:testsets}). In addition, the author performed two further checks in Finnish, which involved 10 accepted and 30 rejected plantings to verify if changes were valid errors of the intended type (Section~\ref{sec:models}), and 20 flagged versus 20 not flagged unmodified Finnish units to check if the Finnish says the same as the English. The sheets used for these checks showed neither the status of a planting nor the verdict of the checker.

\section{Results}
\subsection{Fine-tuning and testing in the standard way}\label{sec:standard}

The study first examined the standard way, where the model was fine-tuned on the training manuals and tested on the unmodified text of the two test manuals, as shown in Table~\ref{tab:standard}. For the fine-tuned back-translator, COMET rose from 80.0 to 84.9, and chrF rose from 50.9 to 70.7. However, on new text, COMET rose only from 76.9 to 77.6, while chrF rose only from 48.5 to 50.6. In addition, on text close to the training text, chrF rose from 52.6 to 83.0. For new text, Clef judged 80.5~\% of the untuned back-translations to say the same as the English version, whereas only 67.8~\% of the fine-tuned ones did so. Word-for-word copies of the English version were found in 1.3~\% of untuned back-translations and in 21.1~\% of fine-tuned back-translations. Therefore, most of the gain of approach A came from text that the back-translator had seen in training.

Regarding approach B on the unmodified test text, the untuned checker flagged 18.5~\% of the unmodified units and 32.2~\% of the harmless edits. In this case, only 47.0~\% of its answers followed the required form. In contrast, the fine-tuned checker flagged 3.4~\% of the unmodified units and 12.7~\% of the harmless edits, and every answer followed the required form. Fine-tuning made the checker of approach B both quieter and more consistent in the form of its answers on the same unmodified text.

\begin{table}[t]
\centering
\caption{Standard evaluation on the unmodified text of the two test manuals using the untuned and fine-tuned model, where fine-tuned values show the mean of three seeds and the range of the seeds.}
\label{tab:standard}
\footnotesize
\setlength{\tabcolsep}{3pt}
\begin{tabularx}{\linewidth}{>{\raggedright\arraybackslash}X r r l}
\toprule
 & Untuned & Fine-tuned & \scriptsize Seed range \\
\midrule
\multicolumn{4}{l}{\textit{A: back-translation of the unmodified test text}} \\
COMET, all units & 80.0 & 84.9 & \scriptsize 84.8 to 84.9 \\
COMET, new text & 76.9 & 77.6 & \scriptsize 76.7 to 78.2 \\
chrF, all units & 50.9 & 70.7 & \scriptsize 70.1 to 71.1 \\
chrF, text close to training text & 52.6 & 83.0 & \scriptsize 81.1 to 84.7 \\
chrF, new text & 48.5 & 50.6 & \scriptsize 48.9 to 51.8 \\
same meaning as the English version, new text (\%) & 80.5 & 67.8 & \scriptsize 64.6 to 69.7 \\
word-for-word copy of the English version (\%) & 1.3 & 21.1 & \scriptsize 17.3 to 24.6 \\
alerts on unmodified text, element rules (\%) & 24.1 & 17.5 & \scriptsize 16.6 to 17.9 \\
false alarms on meaning-preserving edits, element rules (\%) & 25.9 & 32.2 & \scriptsize 29.9 to 33.3 \\
\midrule \multicolumn{4}{l}{\textit{B: verdict on the unmodified test text}} \\
alerts on unmodified text (\%) & 18.5 & 3.4 & \scriptsize 2.5 to 4.1 \\
false alarms on meaning-preserving edits (\%) & 32.2 & 12.7 & \scriptsize 9.7 to 16.9 \\
answers in the required form (\%) & 47.0 & 100.0 &  \\
\bottomrule
\end{tabularx}
\end{table}

\subsection{Planted errors: untuned versus fine-tuned}\label{sec:planted}

\begin{figure*}[t]
\centering
\includegraphics[width=\textwidth]{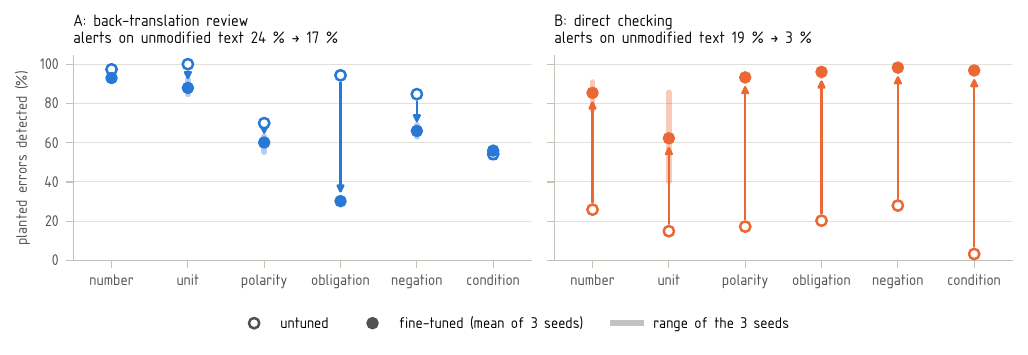}
\caption{Detection per error type from the untuned model, represented by open circles, to the fine-tuned model, represented by filled circles with a mean of three seeds and a range band, showing approach A on the left and approach B on the right with alerts on unmodified text in the panel titles.}
\label{fig:headline}
\end{figure*}

The planted errors show the main result of the study, where training moved the two approaches in opposite directions, as seen in Table~\ref{tab:headline} and Fig.~\ref{fig:headline}. Without the conditions, detection in approach A fell from 87~\% to 67~\%, whereas in approach B it rose from 22~\% to 90~\% (Table~\ref{tab:headline}). With conditions, the share of detected planted errors in approach A fell from 81~\% to 65~\%, which is most evident for weakened obligations that dropped from 94~\% untuned to 30~\% fine-tuned. In addition, detection of negation fell from 85~\% to 66~\%, and detection of units fell from 100~\% to 88~\%. In contrast, the share of detected planted errors in approach B rose from 19~\% to 91~\%. The four word-level types rose to 93~\% to 98~\%, where polarity increased from 17~\% to 93~\%, obligation from 20~\% to 96~\%, negation from 28~\% to 98~\%, and condition from 3~\% to 97~\%. Moreover, detection of numbers rose from 26~\% to 85~\%, and detection of units rose from 15~\% to 62~\%. Alerts on unmodified text were reduced from 24~\% to 17~\% for approach A and from 19~\% to 3~\% for approach B.

The three seeds agreed closely for most types, although they spread widely for the units of approach B, which ranged from 40~\% to 86~\%, and for its numbers, which ranged from 77~\% to 91~\%. For the error types without conditions, fine-tuning resulted in a change of -20 points for approach A, with a 95~\% interval from -24 to -17, and +68 points for approach B, with an interval from +65 to +71, based on resampling whole English texts. In addition, the seeds alone ranged from -22 to -18 points for approach A and from +64 to +72 points for approach B. Nevertheless, Clef as the decision of approach A showed almost no change by fine-tuning, with an increase of 4 points. Section~\ref{sec:hides} explains why this occurs.

\begin{table}[t]
\centering
\caption{Detection of planted errors per error type, alerts and false alarms in percent for the untuned and fine-tuned model, based on the mean of three seeds across 24 languages, with approach A decided by the element rules.}
\label{tab:headline}
\footnotesize
\begin{tabularx}{\linewidth}{>{\raggedright\arraybackslash}X rrrr}
\toprule
 & \multicolumn{2}{c}{A (\%)} & \multicolumn{2}{c}{B (\%)} \\
\cmidrule(lr){2-3}\cmidrule(lr){4-5}
Error type & untuned & fine-tuned & untuned & fine-tuned \\
\midrule
number & 97 & 93 & 26 & 85 \\
unit & 100 & 88 & 15 & 62 \\
polarity & 70 & 60 & 17 & 93 \\
obligation & 94 & 30 & 20 & 96 \\
negation & 85 & 66 & 28 & 98 \\
condition & 54 & 56 & 3 & 97 \\
\midrule
all planted errors & 81 & 65 & 19 & 91 \\
all without conditions & 87 & 67 & 22 & 90 \\
alerts, unmodified text & 24 & 17 & 19 & 3 \\
false alarms, harmless edits & 26 & 32 & 32 & 13 \\
\bottomrule
\end{tabularx}
\end{table}

\begin{figure}[t]
\centering
\includegraphics[width=\linewidth]{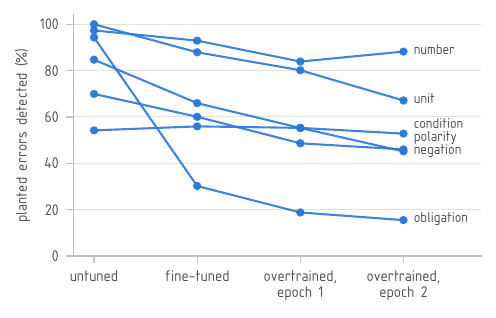}
\caption{Share of planted errors detected in the back-translation for approach A, comparing the untuned, fine-tuned, and overtrained models after epoch 1 and epoch 2.}
\label{fig:intensity}
\end{figure}

\subsection{Approach A: what fine-tuning hides}\label{sec:hides}

An example from the test set~\cite{KV.SeederDA} shows that the English version reads ``Used oil must be collected and disposed of.'', while the Estonian version received a weakened obligation, using ``should'' instead of ``must''. The fine-tuned back-translator returned ``Used oil must be collected and sent to a disposal company.'', which means the planted error disappeared. For weakened obligations, fine-tuning changed the outcomes, where carried results dropped from 94~\% to 30~\%, hidden ones rose from 3~\% to 62~\%, and garbled results increased from 3~\% to 8~\%, as the back-translator restored the expected instruction rather than corrupting the sentence. In addition, more training hid more errors, as the overtrained models detected 55~\% of all planted errors and 17~\% of the weakened obligations (Fig.~\ref{fig:intensity}). Numbers stayed visible in most cases at every training intensity. A pre-registered test confirmed in every seed that fine-tuning raised hiding more in text close to the training text than in new text, where 71~\% to 73~\% of close units were hidden versus 37~\% to 49~\% of new units for weakened obligations. The interaction of training and similarity in the logistic model had log-odds from 0.83 to 0.94 across the three seeds, and all its 95~\% intervals lay above zero, including when the bootstrap resampled whole English texts instead of pages, with a lowest bound of 0.21. In addition, the untuned back-translator hid only 3~\% of the weakened obligations in new text, which means hiding is not only copying of training text.

The author's check of 117 back-translations agreed with the element rules in 85~\% of the items, whereas it agreed with Clef in 66~\%. Regarding weakened obligations, the author saw the planted error in 9 of 9 untuned back-translations, but in 0 of 9 fine-tuned ones. Clef judged most pairs of ``must'' and ``should'' as the same meaning, with an agreement with the author of 44~\%, and therefore Clef missed this hiding (Section~\ref{sec:planted}). For removed conditions, the author saw the removal in 19 of 19 back-translations, while the element rules agreed in 42~\% and Clef in 32~\%. As a result, the study does not interpret the condition result of approach A.

Moreover, 7 of 117 items could not be judged, because the back-translation changed the meaning of the whole sentence, not only the planted point. Such items would go to a translator in practice, which costs exactly the work that approach A should save. The fine-tuned back-translator restored what the operator manuals usually say.

\begin{figure}[t]
\centering
\includegraphics[width=\linewidth]{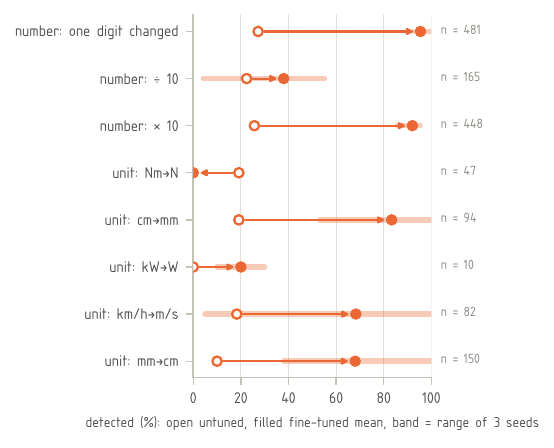}
\caption{Detection of the number and unit changes of the test set for approach B, which differ from the changes in training, showing the untuned model as open circles and the fine-tuned mean as filled circles with a range band of three seeds and the number of test items n.}
\label{fig:bchange}
\end{figure}

\begin{figure*}[t]
\centering
\includegraphics[width=0.9\textwidth]{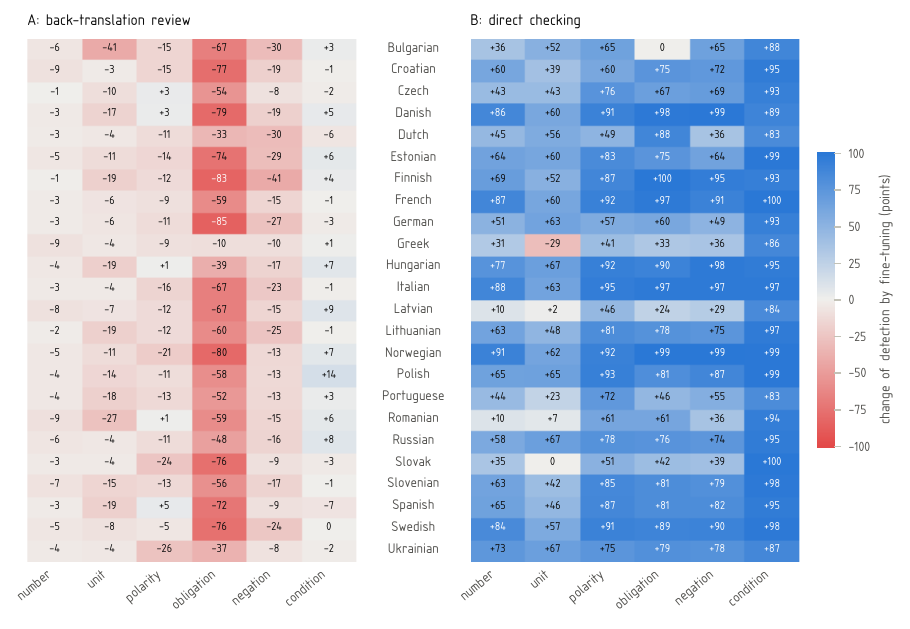}
\caption{Change of detection by fine-tuning per language and error type in points, calculated as the mean of three seeds minus the untuned model, where blue indicates more errors detected and red indicates fewer, showing approach A on the left and approach B on the right. The changes were computed before rounding. Some cells are small, with as few as 3 items for a word-level type and 8 for units in one language.}
\label{fig:languages}
\end{figure*}

\begin{figure*}[t]
\centering
\includegraphics[width=\textwidth]{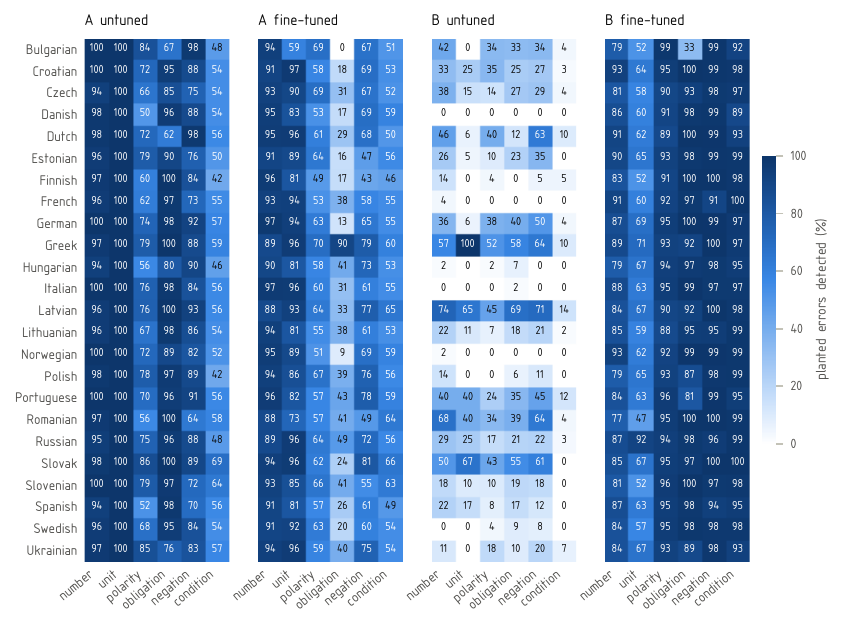}
\caption{Detection of planted errors per language and error type in percent for the untuned and fine-tuned model, based on the mean of three seeds for approach A and approach B. Some cells are small, with as few as 3 items for a word-level type and 8 for units in one language.}
\label{fig:languagesabs}
\end{figure*}

\subsection{Approach B: what the checker learns}\label{sec:learns}

The untuned checker could not do the task in either form of prompt. In the absence of examples, 52~\% of the answers of the untuned checker to planted errors lay outside the required form, where most of them repeated the input or stated the same meaning in the target language. A lenient reading of these answers changed the gain of fine-tuning only from +68 to +66 points for error types without conditions, which shows that the answer form does not explain the gain. Without examples, the untuned checker detected 19~\% of the planted errors at 19~\% alerts, a result that is close to chance. In contrast, the untuned checker flagged almost everything with 99~\% alerts when it received three examples. The fine-tuned checker detected 93~\% to 98~\% of the word-level errors, where gpt-oss-120b planted the training errors and Gemma 4 (26B) planted the test errors. In addition, the answers of the fine-tuned checker named the error and quoted the values. The quoted values were correct in 92~\% to 99~\% of the flagged polarity errors, 69~\% to 91~\% of the numbers, and 39~\% to 74~\% of the units. The untuned checker never quoted the values correctly.

Numbers and units transferred only partly from the training changes to the different changes of the test, as shown in Fig.~\ref{fig:bchange}. For example, a changed digit was detected in 93~\% to 99~\% of cases, and a number ten times larger was detected in 86~\% to 95~\%. However, a number ten times smaller was detected only in 4~\% to 55~\% of cases. Moreover, ``Nm'' written as ``N'' was never detected, resulting in 0~\% in every seed, while ``kW'' written as ``W'' was detected in 10~\% to 30~\% of cases. The checker learned the kinds of changes it saw in training more than a general comparison of numbers and units.

The fine-tuned checker flagged 20 of the 500 unmodified Finnish units. The 500 units are all unmodified Finnish units of the rake manual, and the checker of seed 1 checked all 500. In addition, the 20 unflagged units of the check were drawn at random from the other 480 units. The alerts of the rule check on these Finnish units were not part of the check. In a Finnish check, only one of these 20 alerts showed a real difference, which was a missing ``whenever possible'', while the other 19 shared the same meaning. These false alerts were triggered by exactly two surface cues, namely the normal Finnish obligation ``tulee'' read as ``should'' in 10 alerts and a negation reported missing where the Finnish used a positive or restrictive form in 9 alerts. The Finnish translator often restructured sentences, sometimes more compactly, with the same meaning, which made some items hard to judge. Of the 20 unflagged units, two differed, where one contained an added word and one had an alignment error of the corpus. Therefore, the detection rates of planted errors do not carry over to the precision of alerts on real translations, at least in Finnish.

The rule check alone detected every planted number and unit error, which produced alerts on 0.8~\% of the unmodified units versus false alarms on 38~\% of the meaning-preserving edits, including every thousands separator and every unit spelling such as N·m. In addition, the checker alone detected 90~\% of the errors without conditions, with 3~\% alerts and 13~\% false alarms. The combination detected 97~\% of these errors, with 4~\% alerts and 39~\% false alarms, as shown in Table~\ref{tab:rules}.

\begin{table}[t]
\centering
\caption{Detection of planted errors per error type, alerts on unmodified text, and false alarms on meaning-preserving edits in percent, for the rule check alone, the fine-tuned checker alone (mean of three seeds), and both combined, across 24 languages.}
\label{tab:rules}
\footnotesize
\setlength{\tabcolsep}{3pt}
\begin{tabularx}{\linewidth}{>{\raggedright\arraybackslash}X rrrrr}
\toprule
 & number & unit & without conditions & alerts & false alarms \\
\midrule
rule check & 100 & 100 & 35 & 0.8 & 37.7 \\
checker & 85 & 62 & 90 & 3.4 & 12.7 \\
rule check + checker & 100 & 100 & 97 & 4.0 & 38.7 \\
\bottomrule
\end{tabularx}
\end{table}

\subsection{Copied and new text}\label{sec:copied}

The fine-tuned checker detected 88~\% to 93~\% of the errors without conditions in every similarity group, which shows no loss on copied text in Table~\ref{tab:similarity}. However, approach A lost in every group, where exact copies dropped from 85~\% to 61~\% whereas new text dropped from 90~\% to 75~\%. For weakened obligations, the performance fell from 94~\% to 26~\% in exact copies versus from 94~\% to 42~\% in new text. In addition, the alerts of the fine-tuned checker remained between 2~\% and 5~\% in every group.

\begin{table}[t]
\centering
\caption{Detection of planted errors without conditions in percent, per similarity group of the test text to the training text, for untuned and fine-tuned (mean of three seeds) versions of approach A and approach B.}
\label{tab:similarity}
\footnotesize
\setlength{\tabcolsep}{3pt}
\begin{tabularx}{\linewidth}{>{\raggedright\arraybackslash}X rrrrr}
\toprule
 & & \multicolumn{2}{c}{A (\%)} & \multicolumn{2}{c}{B (\%)} \\
\cmidrule(lr){3-4}\cmidrule(lr){5-6}
Similarity group & $n$ & untuned & fine-tuned & untuned & fine-tuned \\
\midrule
exact copy & 1243 & 85 & 61 & 24 & 93 \\
high (at least 0.8) & 662 & 87 & 58 & 26 & 92 \\
partial (0.5 to 0.8) & 689 & 83 & 65 & 19 & 88 \\
new text & 1725 & 90 & 75 & 21 & 89 \\
\bottomrule
\end{tabularx}
\end{table}

\subsection{Languages}\label{sec:languages}

The opposite directions of the two approaches appeared in all 24 languages, as shown in Fig.~\ref{fig:languages}. In every language, approach A lost weakened obligations by 10 to 85 points, whereas approach B gained in almost every language and error type. However, approach B showed no change for Slovak units and Bulgarian obligation with 3 items and 29 points fewer for Greek units. In addition, the smallest positive gains of approach B were observed for Latvian and Romanian numbers and units, where the gains ranged from 2 to 10 points.

The untuned checker behaved very differently by language, as illustrated in Fig.~\ref{fig:languagesabs}. In Danish, French, Italian, Norwegian, Swedish, and Hungarian, the untuned checker detected 0~\% to 9~\% of the planted errors and answered ``Same meaning.'' to almost every item. In contrast, it flagged up to 74~\% of the planted errors in Latvian. The fine-tuned checker reached 81~\% to 100~\% for the word-level types in every language, apart from Bulgarian obligation with 3 items. Units were the weakest type of the fine-tuned checker in every language, with values between 47~\% and 92~\%.

\section{Discussion}
\subsection{Why the two training workflows act in opposite directions}\label{sec:why}

In approach A, the back-translator learns to produce the English version of the training manuals. Each English sentence is the target of about 20 training pairs, and the fine-tuned back-translator reproduced the memorised English side of these multi-parallel data most of all in text close to the training text (Section~\ref{sec:hides}), similarly as in~\cite{Aharoni.2019}. In the training pairs of this study, a faithful back-translation of the other language version lacks an obligation such as ``must'' in 40~\% of the pairs where the English version contains it. Many of these differences are style, for example an imperative or an infinitive where the English writes ``must be checked''. These training pairs teach that English instructions take ``must'' whatever the other language version says, so the fine-tuned back-translator restores ``must'' where the translation says ``should''. In this study, hiding grew with the training intensity (Fig.~\ref{fig:intensity}), similarly as memorisation grows with the number of repetitions of a training example in~\cite{Carlini.2023,Raunak.2022a}. The back-translator repaired planted errors on their way back, similarly as reported for general systems~\cite{Somers.2005,Moon.2020,Agrawal.2022}, and fine-tuning on the manufacturer's own manuals strengthened this repair. The results support the pre-registered hypothesis, which states that fine-tuning raises hiding more in text close to the training text (Section~\ref{sec:hides}). In addition, hiding occurred also on new text, which remained far above the level of the untuned model. The memory of single training sentences and learned conventions of the operator manuals provide two proposed explanations, although this study did not isolate their separate contributions.

In approach B, the checker learns the difference itself as its target, not the English text. In short, the checker workflow rewards reporting a difference, whereas the back-translator workflow rewards restoring the expected English text. The checker generalised to word-level errors of another planter, but it transferred numbers and units only partly. The two training workflows differ in training target, data, supervision, and selection, which means the design does not isolate one cause. However, the difference in training target, restoring the English text versus reporting a difference, remains a plausible explanation of the opposite directions. In addition, the checker learned from correct pairs that left out stylistic differences in an element, whereas the back-translator learned from all pairs and never saw an error kept in its English target. Both differences likely contributed to the opposite directions, and training both workflows on the same filtered data, including pairs that keep an error for the back-translator, represents a next step.

\subsection{What this means for checking manuals}\label{sec:meaning}

A proposed workflow, which was not tested in this study, involves the checker reading every unit, while a competent translator receives every alert and every answer outside the required form. Unflagged units still need the review required by the standards, because the checker misses some errors. In addition, a reviewer may come to trust the ``Same meaning.'' label and skim unflagged units. In addition, an alert rate of 3~\% is not a measured reduction of the review work, as most alerts in Finnish were false (Section~\ref{sec:learns}) and each false alert costs review time. Human translators restructure sentences with the same meaning, whereas the checker was trained on pairs where a planted change is the only difference. The checker learned language-specific forms of obligation and negation as error cues, and the filter of its correct training pairs, which had removed such forms, likely contributed to this. In addition, keeping such forms as ``Same meaning.'' examples in every language is a next step. A simple rule check comparing numbers and units detected every planted number and unit change similarly as in~\cite{Raunak.2022}. However, it alerted on harmless rewrites such as thousands separators, meaning its normalisation of numbers and units needs care in every language. In addition, the rule check and the checker together covered the weakest types of the checker (Section~\ref{sec:learns}). In contrast, approach A with a fine-tuned back-translator hides exactly the errors the reviewer looks for. In addition, pooled detection rates hide weak types, where the checker never detected a tightening torque in Nm changed to N, and its seeds spread from 40~\% to 86~\% for units. The study where back-translation helped most used a general back-translator~\cite{Ki.2025}. Therefore, the benefit of back-translation for a reader who knows only English depends on a faithful back-translator, while fine-tuning on the same manuals takes this faithfulness away. User studies found that back-translation raises the confidence of users more than the quality of their translations~\cite{Zouhar.2021}, and that users may trust fluent but wrong output~\cite{Martindale.2018}. The check tests consistency with the English reference instead of engineering correctness, which means an unsafe instruction translated faithfully passes the check. The fine-tuned checker could support the review by competent persons; it does not replace it. Similarly, an earlier study proposed that an LLM assistant for operators guides the operator to the source text of the operator manual instead of acting as the final authority~\cite{Oksanen.2026}. In the same way, the checker directs the attention of the reviewer to units, and the reviewer makes the decision.

A human translator may also add or change content on purpose, for example words for better readability in the local language, or a sentence after feedback from that market in a later revision, possibly agreed with the manufacturer. Such a deviation is not a wrong translation, whereas a checker that compares two texts cannot tell an intended deviation from an unintended one. Therefore, this decision needs knowledge of the translation process from a person or from a further AI component, and some work remains for the human reviewer, whose attention the checker can direct. A hypothesis for a next step, not tested in this study, is that translations by a large LLM may stay closer to the structure of the English version and cause fewer false alerts. However, such translations may contain other kinds of errors, such as omissions and added content, and the same checker on LLM translations of the same manuals could test this.

The explanations of the checker quoted wrong values in some cases, for example 39~\% to 74~\% correct for units. In addition, the answers of the checker provide explicit feedback of the kind of LLM explanations; in a user study, such explanations led to more over-reliance than back-translation (13.3~\% against 7.2~\%)~\cite{Ki.2025}, so a reviewer needs correct explanations. Nevertheless, this proof of concept indicates that such support might be possible within operator manuals of agricultural machinery. Firmer conclusions need a far more comprehensive corpus with much more data. Moreover, the proper size of the model remains open, whether it should be smaller than, similar to or larger than the 1.7 billion parameters used here. The EuroLLM family already offers models with 9 and 22 billion parameters~\cite{Martins.2025,Ramos.2026}. However, another model family than EuroLLM may also be a better base for fine-tuning, and this question stays open. For example, the untuned EuroLLM-9B scored similar to or below other open models of similar size in a study that summarised German court decisions~\cite{Nagl.2025}.

\subsection{The English version as reference}\label{sec:english}

This study treated the English version as the reference for every check. It is not known whether English was the source language of each operator manual. The source language may depend on the machine type, for example, Dutch, Norwegian, German or English. We assume that this corpus has no single source language for the whole set of manuals. In addition, it is not known whether all translations come directly from one source. A chain of translations, which functions like a tree, is possible as well. A speculative example of this structure is a set of paths from Dutch to German, Dutch to Norwegian to Swedish to Danish, or Dutch to English to Finnish. It is a general situation for any agricultural machinery manufacturer that translation processes vary highly. 

The Machinery Directive distinguishes ``Original instructions'' from a ``Translation of the original instructions''~\cite{EU.2006}. IEC/IEEE 82079-1 describes translation from the original language into a target language by competent translators~\cite{IEC.2019}. A difference between the English version and another language version may come from the translation path and not only from an error. The English version remains a practical reference for one expert who reads English.

\subsection{Languages in tuning}\label{sec:langtuning}

This study fine-tuned one model on all 24 language versions of the operator manuals together, where the opposite directions of approach A and approach B appeared in every language (Section~\ref{sec:languages}). However, the earlier studies that fine-tuned EuroLLM, which are described in Section~\ref{sec:sota-ft}~\cite{Zhu.2025,Zhang.2025,Guttmann.2025}, cover at most eight language pairs. None of the studies found compares a set of 24 European languages.

\subsection{Building the corpus}\label{sec:corpusbuild}

The corpus was more laborious to build than expected. Earlier studies with operator manuals of the same manufacturer found an identical page layout across three and six language versions~\cite{Gun.2025,Oksanen.2026}, which made page-based alignment look simple. However, single translations inserted or omitted pages in the 12 manuals of this study, and one manual reflowed its text in every language. As a result, every language version needed its own page map, and that specific manual was left out. In addition, the automatic validation of the alignment needed calibration by the author and judged only part of the pairs, so alignment errors remain, as the Finnish check found one among 40 units (Section~\ref{sec:learns}). This experience shows that a study of this kind should reserve sufficient time for the preparation of the data.

\subsection{Limitations}\label{sec:limitations}

The planting of a negation is especially difficult in complex sentences. For example, the test set contains a self-contradiction made by removing a negation, where ``the maximum permitted speed of 540 rpm may be exceeded'' would become a plausible error if it read ``the speed of 540 rpm may be exceeded''. In addition, the test set contains implausible instructions made in the same way, such as ``transport people on the machine''. In languages with negative concord, for example Czech and Slovenian, removing ``ne'' leaves a word for ``nobody'' or ``no'', which means the meaning stays negative. As a result, larger datasets need care in every language, because the changed sentence has to be plausible and not only changed.

The test sets contain planted errors only, and they contain no errors made by translators. Moreover, all operator manuals come from one manufacturer, and the study uses one model of 1.7 billion parameters. English is the only reference language of the study, and the study checks no pairs between two other languages. In addition, every test manual describes a new model of a machine type that also appears in the training manuals, while a machine type absent from training was not tested. The automatic decision of approach A is imperfect, as neither the element rules nor Clef measure removed conditions reliably. Furthermore, the study replaced the reviewer with an automatic decision and did not measure how a reviewer relies on the flags and explanations of the checker. A user study with reviewers of operator manuals, as done for other texts~\cite{Ki.2025}, represents a next step. The test manuals overlap with the training text, where 69~\% and 42~\% of their text units appear word for word in the training manuals. The results show the direction and the size of the effects within this setting, but they do not show general performance figures. In addition, the checks by the author cover Finnish and English only, which means no bilingual check of the planted errors or of the alerts in the other 23 languages was performed. The validation of the plantings favoured short changes in one place, while long conditions and constructions of several words remained underrepresented in the test sets. Moreover, the test sets cover sentences and paragraphs only, whereas headings, lists, tables, captions and text in figures are not covered. Clef filtered the planted errors of the test sets and served as the alternative judge of approach A. However, the element rules made the main decision of approach A, while Clef judged only the removed conditions, whose result is not interpreted. In addition, no pretrained quality-estimation model, such as xCOMET, was compared as a baseline, as the paper compares two training workflows of one LLM, not systems.

\section{Conclusion}
Current standards leave the review of every translation to competent persons, and this review process is the bottleneck of multilingual operator manuals. This proof of concept evaluated, within operator manuals of agricultural machinery, whether a fine-tuned LLM keeps visible or detects discrepancies in translated instructions, which represents a precondition for a possible support of that review, not the support itself. The back-translator trained on the manufacturer's manuals concealed planted errors, especially in text close to the training text, as the pre-registered hypothesis expected. In contrast, the checker trained on the same manuals with synthetic errors detected most of the planted errors, although it remained weak for some unit changes. However, a Finnish check of alerts on real translations showed that most alerts were false, which means that the practical usefulness for naturally written translations remains unestablished, despite the strong detection of planted discrepancies. In addition, the result is in line with earlier observations that back-translation can repair errors on the way back. Overall, the result indicates that a fine-tuned checker might support the review. Firmer conclusions need a far more comprehensive corpus with much more data, and they require an answer to the question of the proper model size.

\begin{ack}
   The author thanks the Kverneland Group for providing the open-access manuals used in this study, and especially Peter van der Vlugt for his support to get insight for these publicly available operator manual documents.

\end{ack}

\makeatletter\def\@bibliosize{\small\baselineskip=0.95\baselineskip}\makeatother
\bibliographystyle{IEEEtran}
\bibliography{literature,manuals}

\end{document}